# A Novel Method for Accurate & Real-time Food Classification: The Synergistic Integration of EfficientNetB7, CBAM, Transfer Learning, and Data Augmentation

Shayan Rokhva, Babak Teimourpour*

Department of Information Technology Engineering, Faculty of Industrial and System Engineering, Tarbiat Modares University, Tehran, Iran

**ABSTRACT:**

Integrating artificial intelligence into modern society is profoundly transformative, significantly enhancing productivity by streamlining various daily tasks. AI-driven recognition systems provide notable advantages in the food sector, including improved nutrient tracking, tackling food waste, and boosting food production and consumption efficiency. Accurate food classification is a crucial initial step in utilizing advanced AI models, as the effectiveness of this process directly influences the success of subsequent operations; therefore, achieving high accuracy at a reasonable speed is essential. Despite existing research efforts, a gap persists in improving performance while ensuring rapid processing times, prompting researchers to pursue cost-effective and precise models. This study addresses this gap by employing the state-of-the-art EfficientNetB7 architecture, enhanced through transfer learning, data augmentation, and the CBAM attention module. This methodology results in a robust model that surpasses previous studies in accuracy while maintaining rapid processing suitable for real-world applications. The Food11 dataset from Kaggle was utilized, comprising 16643 imbalanced images across 11 diverse classes with significant intra-category diversities and inter-category similarities. Furthermore, the proposed methodology, bolstered by various deep learning techniques, consistently achieves an impressive average accuracy of 96.40%. Notably, it can classify over 60 images within one second during inference on unseen data, demonstrating its ability to deliver high accuracy promptly. This underscores its potential for practical applications in accurate food classification and enhancing efficiency in subsequent processes.

**KEYWORDS:**

Deep Learning, AI-driven Food Classification, EfficientNetB7, Convolutional Block Attention Module (CBAM), Transfer Learning, Data Augmentation

## 1) Introduction

### 1.1 The concept and importance of study

The classification, detection, and segmentation of food images using AI and computer vision offer significant potential for innovation and efficiency. These technologies facilitate and automate nutritional tracking, support AI-powered kitchens, and address food waste management (Lubura et al., 2022; Rokhva et al., 2024a). Additionally, with slight modifications, food image recognition techniques can be adapted to monitor crops and other items throughout the supply chain, boosting productivity across food production and consumption processes (Attri et al., 2023; Zhou and Chen, 2023).

Developing accurate and robust models that run promptly on cost-effective hardware is crucial to ensure both effectiveness and efficiency. This has led researchers to explore diverse AI models for better performance, generalization, and speed (Fang et al., 2023; Moumane et al., 2023). In many image analyses and their applications, classification is a critical initial step, as subsequent analyses depend on the accuracy and speed of recognition. Therefore, a model with high accuracy, robustness, and rapid processing of real-world extensive data is pivotal in improving food classification and overall efficiency in the food industry (Chakraborty and Aithal, 2024; Lubura et al., 2022).

Deep learning (DL), a branch of AI, is particularly effective for solving complex problems, especially working with large datasets. Convolutional Neural Networks (CNNs), combining DL with computer vision, are widely applied in visual recognition tasks. Their performance tends to improve as dataset sizes increase, with deeper and more complex models requiring more data (Alzubaidi et al., 2021).

While CNN architectures vary based on their specific use, they consistently follow a structured process involving feature extraction, information summarization, and image classification. CNNs use multiple convolutional layers to extract and combine image features, creating a feature map that aids classification. Max-pooling layers, placed between the convolutional layers, perform feature selection and reduce map size, lowering computational costs. Average pooling, usually at the end, summarizes the most influential and enriched features extracted by previous processes. The final small, data-rich feature map is flattened and passed to the classifiers for prediction (Alzubaidi et al., 2021; Sarraf et al., 2021). Conversely, Artificial Neural Networks (ANNs) require exponentially more parameters for image classification, demanding significant computational resources, making them less practical for such tasks compared to CNNs (Lubura et al., 2022).

Over the past decades, various CNNs have been developed and implemented, including the AlexNet, VGG, GoogleNet, ResNet, MobileNet, and EfficientNet families. Each family contains models that differ in depth, size, and parameters while maintaining a similar structure and underlying concept. However, models among different families also vary in their overall architecture according to specific applications and concepts (Karypidis et al., 2022; Li et al., 2021). For instance, the MobileNet family is designed for quick classification with fewer parameters and computations, making them ideal for mobile and embedded devices. However, they may exhibit lower performance in specific applications (Banoth and Murthy, 2024; Rokhva et al., 2024a). Conversely, the EfficientNet family, developed by Google, employs compound scaling to optimize depth and width, yielding lightweight models for more straightforward and less complex tasks and deeper models for complex applications requiring high accuracy. The enhanced performance of the deeper EfficientNet models has been demonstrated across numerous tasks (Huang et al., 2019; Tan, 2019).

The widespread adoption of these models emphasizes the need to utilize, customize, and improve them for real-world applications while tackling their challenges, such as inferior speed or accuracy, by leveraging DL techniques. (Alzubaidi et al., 2021; Karypidis et al., 2022). For example, while deep and dense models offer higher accuracy using a large number of data, they also lead to more trainable parameters and increased processing times (Tan, 2019). Therefore, leveraging pre-trained knowledge, freezing part of the network, utilizing parallel computation of GPU, and similar techniques can help address this issue (Farahani et al., 2021).

Since we briefly understood the challenges and importance of integrating AI and computer vision into kitchen and food consumption settings, we should also explore their frequent use in modern facilities (Chakraborty and Aithal, 2024). For instance, smart refrigerators utilize hidden cameras and AI models to identify food items and notify users about freshness and expiration dates, thereby aiding in reducing food waste, which constitutes 25-35% of food globally. However, these advanced facilities remain primarily unaffordable for the average consumer, probably due to the expenses of state-of-the-art research, highlighting a significant area for enhancement (Gao et al., 2019; Shweta, 2017).

To improve accessibility for individuals from all walks of life, research must focus on attempts to optimize performance and boost speed while diminishing reliance on costly high-end graphical processing units (GPUs). Additionally, refining the performance and robustness of these systems is crucial (Alijani et al., 2024). Therefore, further exploration and advancements in this field are both necessary and warranted to enhance usability and efficiency.

**1.2 Brief survey of similar works**

This section aims to expand readers' understanding by exploring studies that utilize computer vision and DL techniques in the food industry and related fields. One of the early studies in automatic AI-driven food recognition (Kagaya et al., 2014) employed a CNN architecture on a large food image dataset, emphasizing hyperparameter tuning. Their findings demonstrated that CNNs outperformed Support Vector Machine (SVM), a traditional ML algorithm, using handcrafted features. Similarly, (Kawano and Yanai, 2014) reported that combining features from Deep CNNs with traditional handcrafted image features, such as Fisher Vectors, HoG, and Color patches, led to enhanced food recognition performance, achieving top and top-5% and top-1% accuracies of 72.26% and 92%, respectively, on the UEC-FOOD100 dataset.

(Singla et al., 2016) utilized a Google-Net-based model, achieving impressive accuracies of 99.2% for food and non-food classification and 83.6% for differentiating food categories. (Siddiqi, 2019) applied the VGG16 model, renowned for its numerous trainable parameters, in conjunction with transfer learning, attaining a classification accuracy of 99.27% for AI-driven fruit image recognition. Additionally, (Zhu et al., 2020) introduced a system capable of classifying and detecting multiple food items in refrigerators faster and more accurately than the previous method, boosting the previous F-measure baseline by 3-5%.

Transfer Learning (TL) has become a prominent strategy in DL, enabling the use of existing knowledge to enhance accuracy, accelerate convergence, and improve performance within a decreased number of epochs. Simultaneously, data augmentation can be employed to increase data diversity and quantity, helping to mitigate overfitting and yield robust results. Both techniques effectively address data scarcity. TL enables neural networks (NNs) to be trained with limited data by leveraging prior knowledge, while data augmentation expands and boosts dataset size and variety for better outcomes. The literature consistently supports the idea that combining TL and data augmentation enhances results, highlighting their significance in DL applications (Dhillon and Verma, 2020; Hosna et al., 2022; Iman et al., 2023)

In agriculture, a field close to food recognition, researchers (Kumar et al., 2020) showed that ResNet34 classifies crop images with an impressive accuracy of 99.4%. A recent study (Banoth and Murthy, 2024) used MobileNetV2 for highly accurate soil image classification, exceptionally classifying tests data with 100% accuracy. These DL models have also been widely used for medical purposes. For example, EfficientNetV2 has been applied for chest X-ray and CT image classification, achieving accuracy of 98.33% and 97.48%, respectively (Huang and Liao, 2022). For food classification, researchers (Fakhrou et al., 2021) utilized DCNNs with TL and data augmentation, obtaining an accuracy of 95.55%.

A study (Mazloumian et al., 2020) investigated the classification and segmentation of food waste using DL, TL, and data augmentation, achieving an accuracy of 83.5% for classifying food waste images and a pixel accuracy of 98.5% for segmentation. Another comprehensive study in the food industry (Lubura et al., 2022) implemented a classic CNN model similar to the VGG architecture for food classification, achieving 98.8% performance. This research also estimated food waste at 21.3% by segmenting images from Serbian students' meals over six months in 2022.

Recent advancements in computer vision, such as attention mechanisms, have been used for recognizing food items. Nonetheless, their applications are still limited. Convolutional Block Attention Module (CBAM) (Woo et al., 2018) enhances CNN performance by integrating attention mechanisms across channels and spatial dimensions. By refining feature maps with channel attention to emphasize critical channels, followed by spatial attention to focus on essential regions, CBAM improves model accuracy and feature representation. In one research in the food industry (Hui-jiang et al., 2024) a CBAM-InceptionV3 model, embedding 11 CBAM modules into Inception V3, was proposed to enhance food image classification. Utilizing TL on the Food-101 dataset, the proposed model achieved 82.01% accuracy, outperforming the original InceptionV3 in feature extraction, classification, and training efficiency.

Remarkably, studies focusing on the Food-11 dataset, the dataset used in this research, will be thoroughly examined in the discussion to compare methodologies, results, and limitations.

## 1.3 Research efforts and contributions

Given the significance of AI-driven food recognition in the contemporary world, this study aimed at improving the performance of multiclass food classification using the Food 11 dataset. The critical contribution of the study can be summarized as follows:

1. Acquiring the Food11 dataset from Kaggle, containing 16,643 images across 11 classes, presenting challenges due to data imbalance, intra-class diversity, and inter-class similarities.

2. Leveraging EfficientNetB7, a state-of-the-art and deep model, as a backbone for feature extraction, further enhancing the feature map by Channel and Spatial Attention Modules using CBAM, all done to achieve superior performance.

3. Implementing fine-tuning to improve accuracy and accelerate convergence while diversifying the dataset to mitigate overfitting and enrich the training process by exposing the model to a broader variety of data.

4. Conducting effective hyper-parameter tuning, particularly progressively lowering the learning rate while applying regularization to minimize overfitting and accuracy fluctuations, resulting in stabilized convergence.

5. Surpassing all prior research on the same dataset that utilized a single deep learning model while approaching the performance of ensemble learning methods that combined four DL models.

6. Attaining an evaluation speed that is adequately high for various practical applications despite the model's complexity and extensive parameters.

7. Delivering a comprehensive discussion to evaluate the techniques' effectiveness, analyze results, address challenges, and suggest opportunities for future research enhancements.

### 1.4 Order of study

The remainder of the study is structured as follows: Section 2 outlines the material and methods, Section 3 presents the results and analysis, Section 4 offers an in-depth discussion of the findings, and Section 5 concludes the paper.

## 2) Material and Methods

### 2.1 Overview of research workflow

This research follows a standard workflow in ML, emphasizing model training with a substantial part of the dataset. Performance is optimized by tuning hyperparameters using validation datasets, ensuring that evaluation data remains untouched to prevent leakage and bias. Once optimized, the model is tested only once on unseen test (evaluation) data, with metrics reported accordingly (Al-Alshaikh et al., 2024; Rokhva et al., 2024b). TL and data augmentation are integrated during training and can be adjusted using validation data but not test data.

### 2.2 Hardware configuration

Due to the computational demands of DL and the unstructured nature of image data, which require extensive computations, GPUs are indispensable. However, to make this research more accessible and cost-effective for real-world applications, we opted for the budget-friendly T4 GPU with 16GB of memory instead of the high-performance expensive ones such as A100, striking a balance between efficiency and affordability. Notably, both GPUs are available through Google Colab.

### 2.3 Distribution and challenges of data

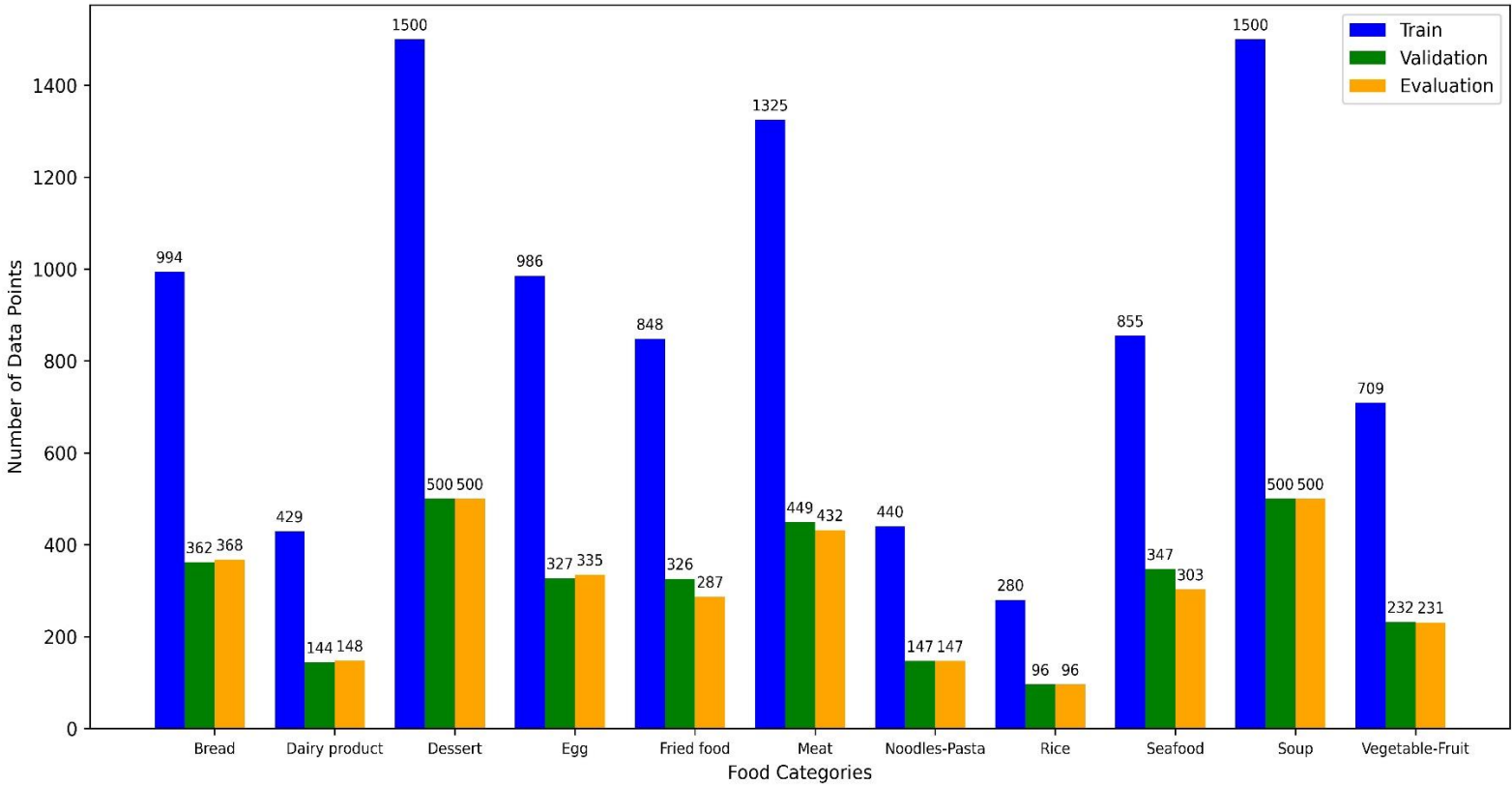

*Figure 1 - Data Distribution*

The Food-11 dataset ("Food-11 image dataset," n.d.), sourced from Kaggle, contains 16,643 images spanning 11 food categories. Figure 1 illustrates the distribution, revealing a notable imbalance that could hinder model performance, especially for underrepresented classes. Approximately 60% of the images are allocated to the training set in each class, with the remainder almost evenly split between validation and evaluation subsets.

The dataset also presents classification challenges due to substantial intra-class diversity and inter-class similarity. Broad categories like desserts and fried food include visually distinct items, complicating accurate classification. Additionally, food categories such as bread, dairy products, desserts, and fried food share similar bright colors such as light brown, white, and yellow, thereby raising the potential for misclassification, which could undermine overall performance.

Additionally, foods such as hamburgers and pizzas, which are typically classified as bread, may initially be confused with fried foods and subsequently with desserts or dairy products. These visual similarities, particularly evident in the first five classes illustrated in Figure 1, from Bread to Fried Foods, heighten the potential for misclassification across categories.

**2.4 Transforming and diversifying data**

The dataset images varied in dimensions. Consequently, they were all resized to 256x256 pixels, a common size in computer vision projects that strikes a balance between resolution and speed. After converting them to numerical values, normalization was applied using the mean and standard deviation (STD) of ImageNet. To enhance performance without increasing the training data, augmentation techniques (Moreno-Barea et al., 2020) like rotation, horizontal flipping, marginal color jittering, and partial erasing were randomly applied to the training set. Consequently, some images remained unchanged, while others were modified by one or more techniques.

This approach increased the diversity of training data, generally leading to a more robust learning process and better model performance since the model was exposed to a broad range of data (Alijani et al., 2024). In this context, Figure 2 showcases a batch of 16 images from the training set following applying the aforementioned techniques. Furthermore, it highlights some of the challenges outlined in Section 2.3.

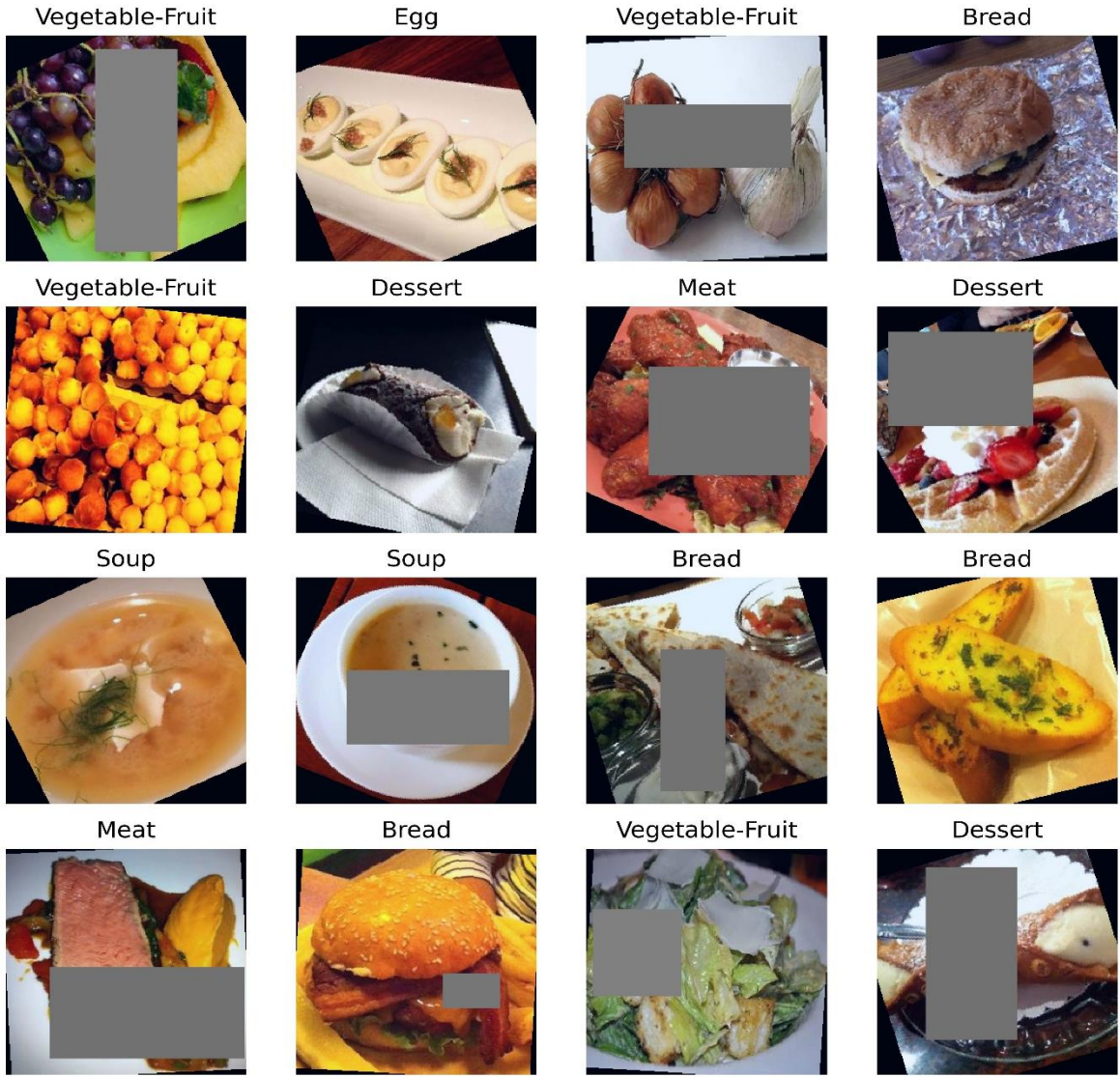

*Figure 2 – A training batch of 16 images after data diversification techniques*

**2.5 EfficientNetB7 backbone**

The EfficientNet family, introduced by Google in 2019 (Tan, 2019), constitutes a state-of-the-art architecture specifically designed to enhance performance efficiently through a compound scaling approach; this methodology simultaneously scales the model's depth, width, and resolution, thereby yielding superior performance compared to the isolated scaling of any single dimension. Encompassing models from B0 to B7, the EfficientNet family includes lightweight variants such as B0, B1, and B2, which possess fewer trainable parameters and consequently require less computational power and time; however, these models often exhibit limitations in their performance on complex tasks. Conversely, the deeper models—specifically B5, B6, and B7—are characterized by increased density and computational intensity, rendering them more suitable for large-scale applications requiring high accuracy.

For enhanced comprehension, Table 1 provides a comprehensive comparison of the performance of EfficientNet models against other prominent computer vision architectures on the 1000-class ImageNet dataset; this comparison underscores the superior performance of EfficientNets, which achieve significantly better results with considerably fewer parameters and FLOPs than comparable models. Such efficiency is attributed to the innovative compound scaling technique employed within this family of architectures, which optimally balances model complexity with performance (Arnandito and Sasongko, 2024; Tan, 2019).

Table 1 – A comprehensive comparison of EfficientNets and other computer vision models

| Model | Top-1 Accuracy | Top-5 Accuracy | Parameters | Parameters/ EfficientNet | FLOPs | FLOPs/ EfficientNet |
|---|---|---|---|---|---|---|
| **EfficientNet-B0** | **77.1%** | **93.3%** | **5.3M** | **1x** | **0.39B** | **1x** |
| ResNet-50 (He et al., 2016) | 76.0% | 93.0% | 26M | 4.9x | 4.1B | 11x |
| DenseNet-169 (Huang et al., 2017) | 76.2% | 93.2% | 14M | 2.6x | 3.5B | 8.9x |
| **EfficientNet-B1** | **79.1%** | **94.4%** | **7.8M** | **1x** | **0.7B** | **1x** |
| ResNet-152 (He et al., 2016) | 77.8% | 93.8% | 60M | 7.6x | 11B | 16x |
| DenseNet-264 (Huang et al., 2017) | 77.9% | 93.9% | 34M | 4.3x | 6.0B | 8.6x |
| Inception-V3 (Szegedy et al., 2016) | 78.8% | 94.4% | 24M | 3.0x | 5.7B | 8.1x |
| Xception (Chollet, 2017) | 79.0% | 94.5% | 23M | 3.0x | 8.4B | 12x |
| **EfficientNet-B2** | **80.1%** | **94.9%** | **9.2M** | **1x** | **1.0B** | **1x** |
| Inception-V4 (Szegedy et al., 2017) | 80.0% | 95.0% | 48M | 5.2x | 13B | 13x |
| Inception-ResNet-V2 (Szegedy et al., 2017) | 80.1% | 95.1% | 56M | 6.1x | 13B | 13x |
| **EfficientNet-B3** | **81.6%** | **95.7%** | **12M** | **1x** | **1.8B** | **1x** |
| ResNeXt-101 (Xie et al., 2017) | 80.9% | 95.6% | 84M | 7.0x | 32B | 18x |
| PolyNet (Zhang et al., 2017) | 81.3% | 95.8% | 92M | 7.7x | 35B | 19x |
| **EfficientNet-B4** | **82.9%** | **96.4%** | **19M** | **1x** | **4.2B** | **1x** |
| SENet (Hu et al., 2018) | 82.7% | 96.2% | 146M | 7.7x | 42B | 10x |
| NASNet-A (Zoph et al., 2018) | 82.7% | 96.2% | 89M | 4.7x | 24B | 5.7x |
| AmoebaNet-A (Real et al., 2019) | 82.8% | 96.1% | 87M | 4.6x | 23B | 5.5x |
| PNASNet (Liu et al., 2018) | 82.9% | 96.2% | 86M | 4.5x | 23B | 6.0x |
| **EfficientNet-B5** | **83.6%** | **96.7%** | **30M** | **1x** | **9.9B** | **1x** |
| AmoebaNet-C (Cubuk et al., 2019) | 83.5% | 96.5% | 155M | 5.2x | 41B | 4.1x |
| **EfficientNet-B6** | **84.0%** | **96.8%** | **43M** | **1x** | **19B** | **1x** |
| **EfficientNet-B7** | **84.3%** | **97.0%** | **66M** | **1x** | **37B** | **1x** |
| GPipe (Huang et al., 2019) | 84.3% | 97.0% | 557M | 8.4x | - | - |

EfficientNetB7, renowned for its exceptional performance in complex tasks (Rokhva et al., 2024c; Tan, 2019), was selected as the backbone for feature extraction. As illustrated in Figure 3, this architecture (Khalil et al., 2022) comprises Mobile Inverted Bottleneck Convolutions (MBConv) organized into seven blocks, beginning with a standard Conv2D layer and progressing through increasingly deeper and wider MBConv layers that extract progressively detailed features. The deeper blocks initially utilize broader 5x5 convolutions before transitioning to 3x3 convolutions, thereby capturing more abstract, high-level information.

Moreover, EfficientNetB7 employs depth-wise separable convolutions within its MBConv layers, significantly reducing computational complexity while preserving feature richness. Each MBConv block consists of an expansion phase, where the input channels are increased to capture intricate details, followed by a compression phase that reduces dimensionality. As the network deepens, later blocks adopt larger kernel sizes and higher expansion ratios, effectively increasing the receptive field; this enhancement enables the model to summarize input data from even small objects more efficiently while extracting complex patterns and abstract representations.

This study sets the input image dimensions at 256x256x3, representing a 256x256 image size across three channels: red, green, and blue (RGB). Consequently, the resulting feature map measures 8x8x2560, showcasing the 8x8-sized feature map enriched with summarized data and a large receptive field. With no modification, the final feature map in EfficientNetB7 is flattened and passed through a fully connected (FC) layer for classification; however, in this study, the feature map undergoes further optimization using the CBAM, which integrates channel and spatial attention mechanisms before reaching the FC layer.

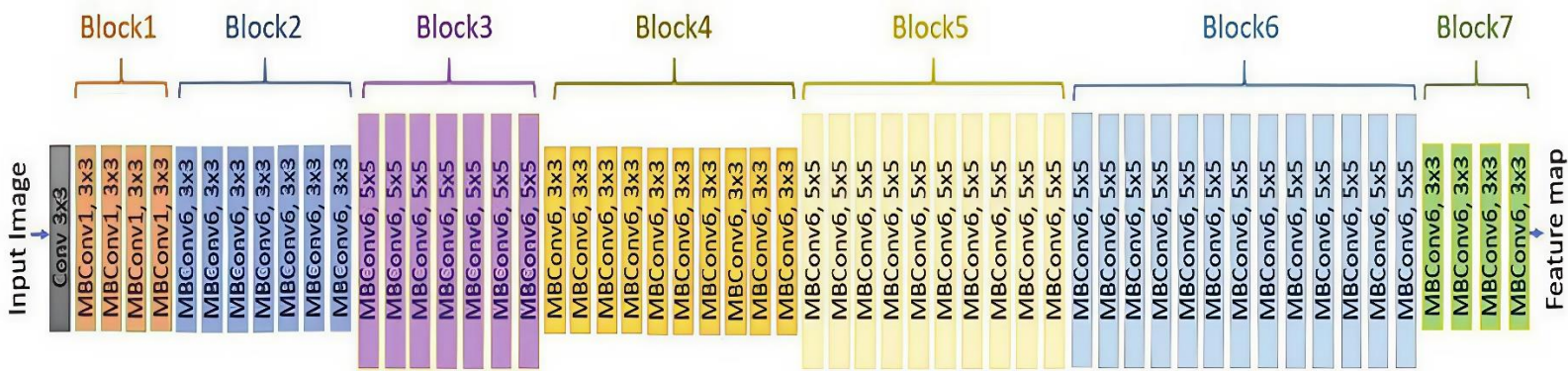

*Figure 3 – EfficientNetB7 backbone structure*

## 2.6 Convolutional Block Attention Module (CBAM)

The Convolutional Block Attention Module (CBAM) is an innovative attention mechanism designed to enhance the performance of CNNs. It does this by refining feature maps through two fundamental processes: focusing on important channels and emphasizing significant spatial regions. CBAM aims to improve the model's ability to recognize and prioritize relevant image features, particularly useful in tasks like image classification and object detection. As CBAM's structure is shown in Figure 4, it consists of a Channel Attention Module (CAM) and a Spatial Attention Module (SAM), operating sequentially, CAM followed by SAM (Woo et al., 2018).

CAM is responsible for evaluating the importance of different channels within the feature map through a three-step process. The first step involves information gathering, where CAM collects global data from the feature map by summarizing values across all spatial locations. This is accomplished using two techniques: global average pooling, which averages the values, and global max pooling, which selects the maximum value. Consequently, these methods yield two distinct representations of channel importance. The second step focuses on weight calculation; here, the two summaries are processed through a small neural network, which learns to assign weights to each channel based on its relevance to the task. Finally, these calculated weights are applied back to the original feature map in the feature enhancement phase. Channels identified as more critical receive higher weights, amplifying their contribution, while less significant channels are diminished.

After refining channel-wise features through CAM, SAM focuses on enhancing specific spatial regions within those channels through a three-step process. The initial step involves extracting contextual information, where SAM analyzes the output from CAM to gather insights about spatial regions by applying both average pooling and max pooling across the channels. This dual approach helps identify which areas of the feature map hold greater significance. The second step is the creation of an attention map; here, the pooled information is combined and processed through a convolutional layer to generate a spatial attention map that highlights critical regions in the feature map for further processing. Finally, in the refinement phase, this spatial attention map is applied to the refined feature map obtained from CAM. Areas identified as essential are emphasized, enabling the model to focus on crucial parts of the input image.

By combining these two processes, CBAM allows deep learning models to dynamically adjust their focus on features that matter most, leading to improved performance in various computer vision applications. This approach enriches feature representation and helps models become more interpretable and efficient in processing visual data. The efficient application of the CBAM module has been proven in numerous studies since its invention. In this study, the extracted feature map from EfficientNetB7, which is rich in data, is modified more through the application of CBAM. Afterward, it is flattened and goes through the FC layer for classification.

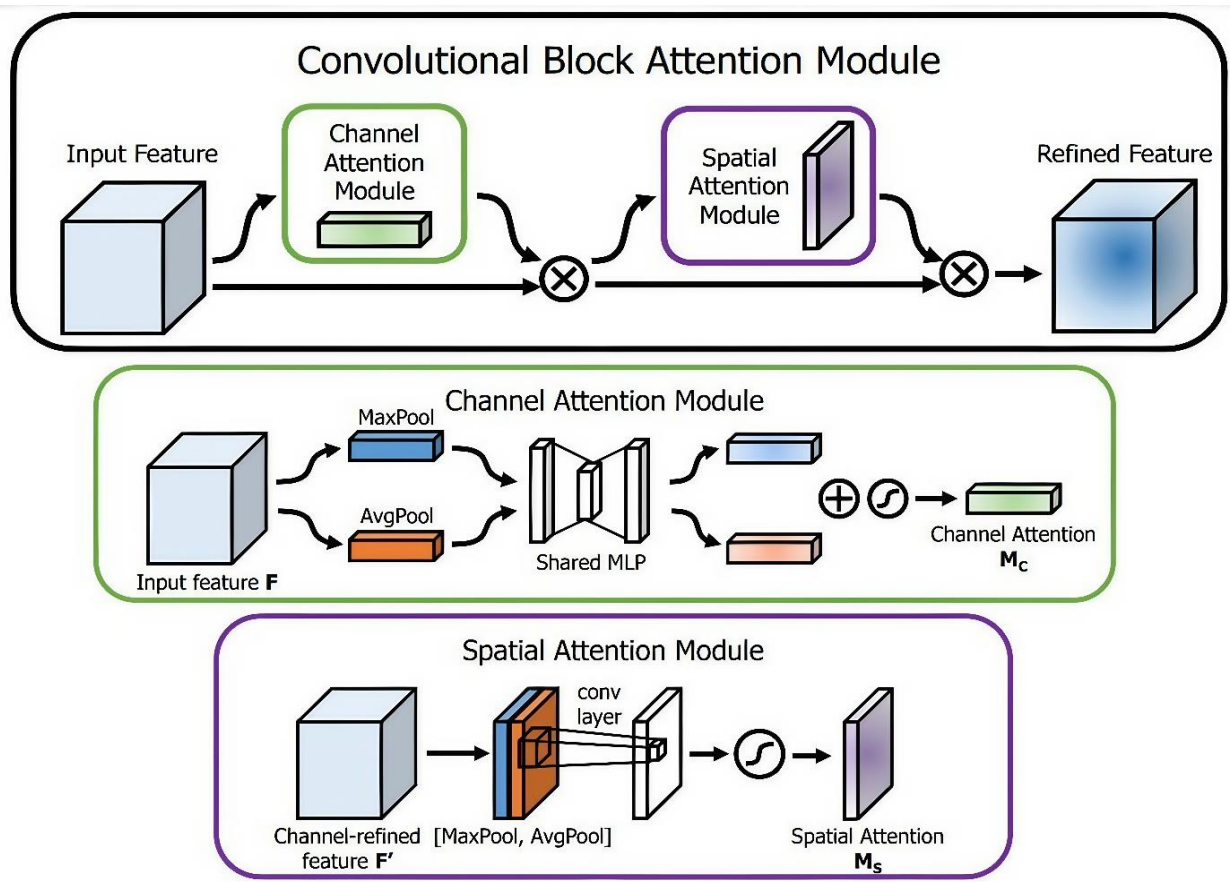

*Figure 4 – Convolutional Block Attention Module (CBAM) and its components*

### 2.7 Transfer Learning

Transfer Learning (TL) involves utilizing the weights and biases from a network trained on one task to enhance performance on a new application, effectively leveraging existing knowledge. This approach offers significant advantages, such as improving initial accuracy, achieving faster convergence, and reaching superior performance within the same number of epochs. In this context, the closer the two applications are, the more effective TL is expected (Hosna et al., 2022).

There are three main types of TL: Feature extraction, where all parameters are frozen except for those in the FC layer, making it suitable for scenarios with limited data or computational resources; Fine-tuning, which allows training of deeper layers responsible for extracting task-specific features while freezing earlier tasked with learning general features; and Full fine-tuning, where all parameters are retrained for additional epochs to adapt them specifically to the new task (Zhuang et al., 2020).

Research indicates that full fine-tuning typically yields better results when sufficient computational resources and large data volumes are available. Even though this method demands more computational power, once the model is optimized, its parameters can be utilized multiple times in inference mode without requiring backpropagation during training. Thus, intensive computation becomes a one-time necessity (Xing et al., 2024; Zhuang et al., 2020).

In this study, the EfficientNetB7 backbone, which aims to extract rich and deep features, was initially imported with the weights and biases of extensive ImageNet data. Yet, to ensure optimal weights that are highly task-specific, all model parameters were exposed to more training epochs.

### 2.8 Loss function, optimizer, and hyper-parameter tuning

Stochastic Gradient Descent (SGD) was employed as the optimizer, while Cross Entropy Loss served as the loss function. The learning rate, the most influential parameter for model training, was set at 0.01. However, to achieve optimal performance, it was not fixed; it halved every five epochs. This strategy allows for accelerated learning in the early stages of training, where faster convergence and larger steps are needed. In comparison, slower learning in later epochs facilitates a better and smoother convergence, leading to the global minimum loss.

L2 Regularization was set to 1-5 to prevent overfitting and enhance generalization. Momentum was configured at 0.9 to accelerate learning and convergence effectively, while the Nesterov accelerator was enabled for improved convergence.

The batch sizes for training, validation, and evaluation were 16, 32, and 32, respectively. The smaller training batch size was chosen due to RAM constraints during backpropagation. It is crucial to report these batch sizes as they can marginally influence average processing speed; larger batch sizes can leverage GPU parallel computations if RAM permits, potentially reducing total computation time.

### 2.9 Evaluating performance and speed

Performance and speed must be reported to compare outcomes with other studies. This study assesses performance using accuracy, a single number indicating the proportion of accurate predictions. Moreover, a confusion matrix was utilized to reveal errors' locations and portions, highlighting classes prone to mutual misclassification. This graphical representation helps us extract insights, guide discussions, and identify improvement areas.

Speed is also critical, varying slightly with batch sizes and differing heavily across GPUs. As mentioned, this study reports training and evaluation speeds using the affordable T4 GPU. Additionally, while both speeds are documented, evaluation speed is essential for real-world applications when the trained model is used in its inference mode.

### 2.10 Generalization and K-fold cross-validation

For improved applicability, results should be generalized through K-fold cross-validation, a standard method for enhancing robustness. Nonetheless, it was not applicable here due to the dataset's initial separation. Hence, to ensure robust results while preventing data leakage or bias, the model was trained, optimized, and then executed on the test data 5 times, each time all parameters were initialized from scratch, and the mean performance from these runs was reported, providing a generalized result that improves robustness similar to 5-fold cross-validation (Kohavi, 1995).

In this scenario, we anticipate five accuracies close to each other while being slightly different due to randomness in data loader creation and applying data augmentation techniques on the training samples. For generalization, the average accuracy should also be close to each of the five individually reported numbers.

## 3) Results and analysis

After multiple trials, the predicted evaluation accuracies on unseen data were consistently close, aligning with our expectations. The five optimal evaluation accuracies obtained were 96.24%, 96.44%, 96.51%, 96.42%, and 96.38%. Based on this, the average reported accuracy of our proposed method is 96.40%. The close proximity of these individual values to the average further confirms the model's robustness. We accept that additional trials may slightly influence the average performance. Nonetheless, the effect would be minimal.

Figure 5 illustrates the smooth and effective convergence of both training and evaluation accuracies, highlighting the benefits of TL in enhancing performance and accelerating convergence. It also demonstrates that data augmentation and regularization helped to prevent overfitting and accuracy fluctuations. Moreover, it shows that the dynamic learning rate improved accuracy, particularly after every five epochs. However, this effect is more pronounced in the earlier epochs, with diminishing returns in later stages.

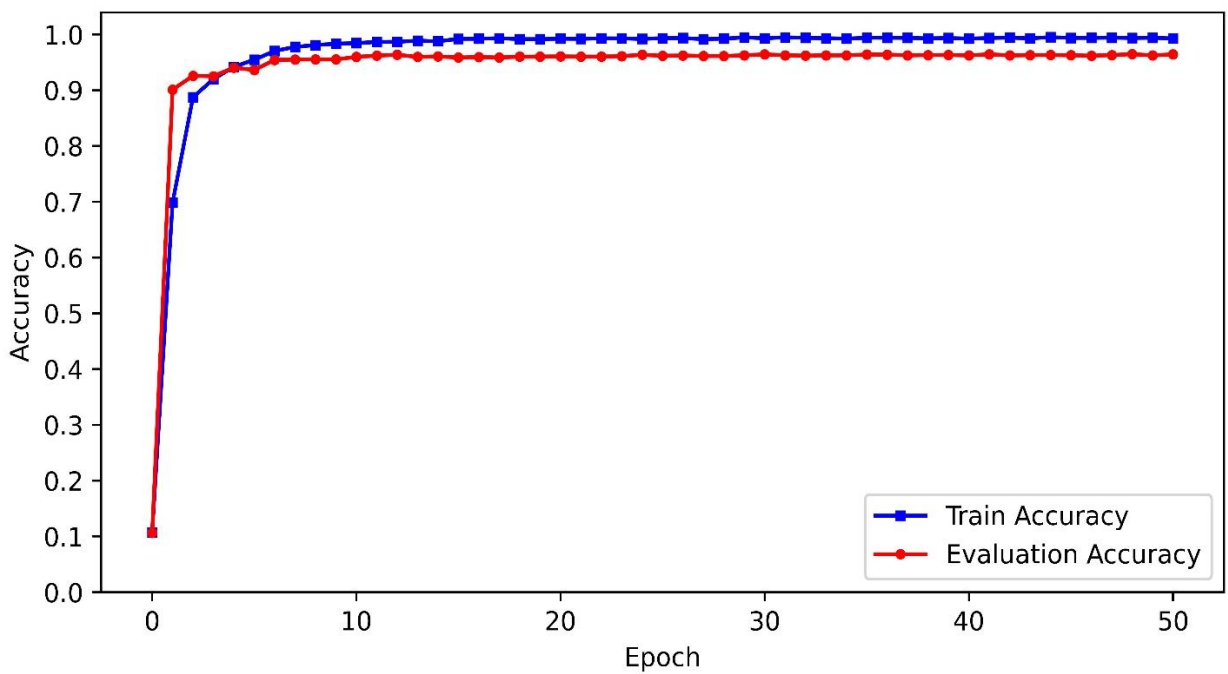

*Figure 5 – Accuracy convergence for train and evaluation*

Figure 6 presents a normalized version of the confusion matrix, ensuring that varying data amounts across food categories do not bias our analysis. The matrix reveals that some specific categories (Noodle-Pasta, Rice, Soup, and Vegetable-Fruits) were classified with nearly perfect accuracy (around or above 99%), while others (Dairy products, Dessert, and Egg) underperformed, with Dairy products showing the lowest performance.

While observing variations in classification performance across categories in ML and DL projects is typical, understanding these discrepancies is crucial for improvement. In this study, we discovered that data imbalance did not heavily influence misclassification. Despite having the least data, "Rice" and "Noodles-Pasta" achieved outstanding performance, comparable to that of "Vegetable-Fruits" and significantly higher than that of "Desserts", which had the most significant proportion of data.

Moreover, "Noodles-Pasta" and "Dairy products" demonstrated starkly different outcomes, with one showing one of the highest accuracies and the other the lowest, despite having the same data volume. Notably, the mentioned items are just instances, and further insights can be drawn by examining and comparing Figures 1 and 6 simultaneously.

The results support our hypothesis from Section 2.3 regarding dataset challenges due to similarities among categories. Food classes with similar colors, such as Bread and Dairy products, underperformed and exhibited mutual misclassifications, which will be further analyzed.

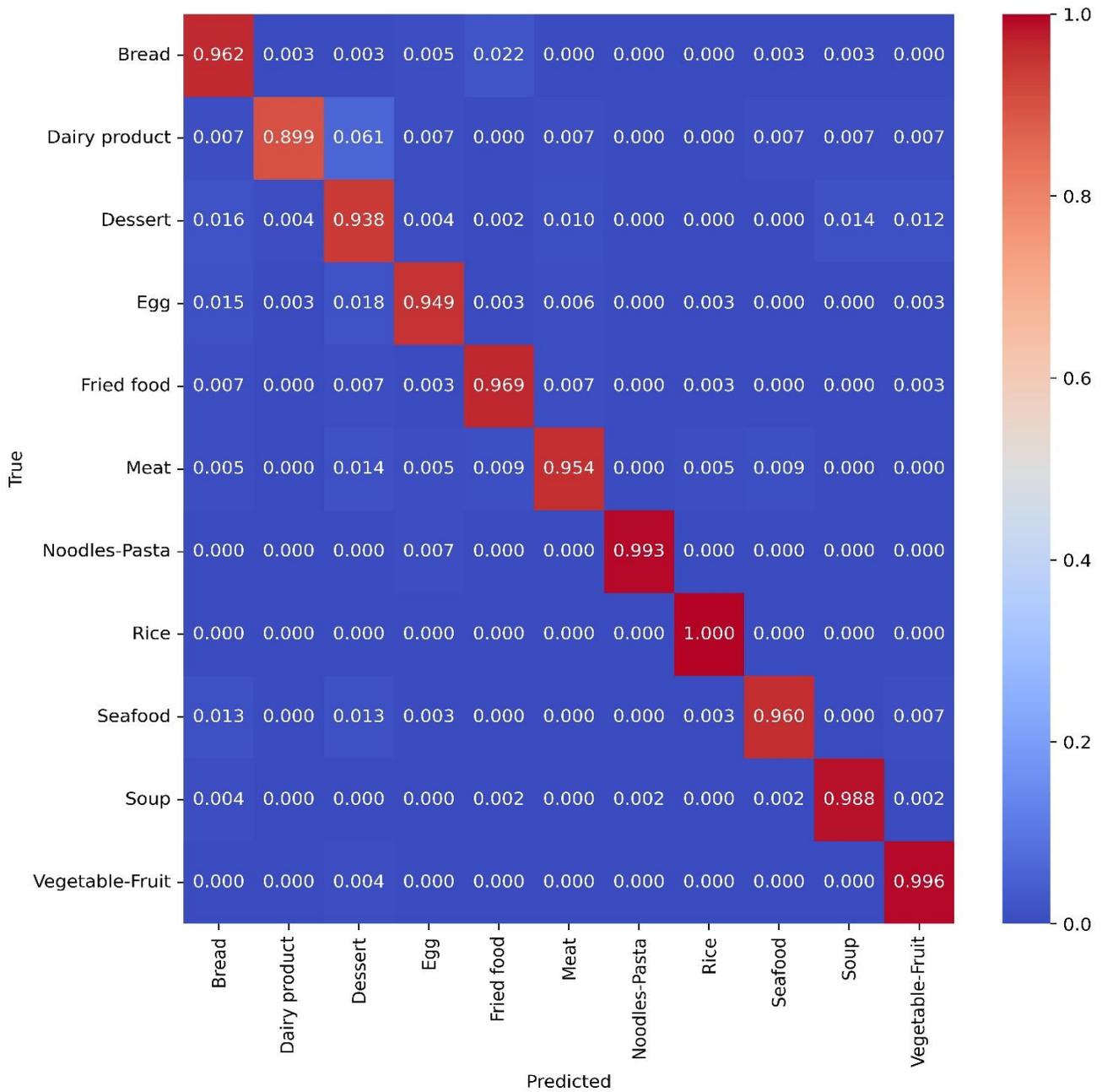

*Figure 6 – Normalized confusion matrix*

In keeping with scientific rigor, it is essential to note that while the reported average accuracy of 96.40% is based on five trials, Figures 5 and 6 reflect the results of a single run. As a result, the values in Figure 5 and Figure 6 may vary slightly between executions. It should be stated that despite potential differences in specific values, the relative rankings of classes, those with the best, average, and worst accuracies, remain relatively constant, supporting the reliability of our analysis and discussion.

Speed was also evaluated. Using a T4 GPU with 16GB memory in Google Colab, training and evaluation speeds were approximately 1.01–1.03 batch/s and 1.95–2.01 batch/s, respectively. With batch sizes of 16 for training and 32 for evaluation, the final speeds reached 16 and 62 images/second. Though speed can vary depending on batch size and hardware, the evaluation speed of 60 images/second on the T4 GPU, paired with high accuracy, is considered suitable for many real-world applications.

## 4) Discussion

### 4.1 Comparison with similar studies

This section examines research efforts that share our focus on accurate Food-11 image classification. In a study (Özsert Yiğit and Özyildirim, 2018), CaffeNet and AlexNet were used for food image recognition, initially obtaining accuracies of 80.51% and 82.07% with the SGD optimizer. When Adam was adopted as the optimizer, the same models achieved enhanced accuracies of 83.7% and 86.92%. The images in their study had a resolution of 512x512 pixels, the highest among similar studies, leading to significantly more computations under the same circumstances.

They (Islam et al., 2018a) conducted a study using a pre-trained InceptionV3 model paired with TL, data augmentation, and images resized to 299x299 pixels, resulting in an accuracy of 92.86%. In a later study (Islam et al., 2018b) a CNN was developed from the ground up. A pre-trained InceptionV3 model was incorporated into the research, with images resized to 224x224 pixels. After employing various DL techniques, the proposed approach achieved an accuracy of 74.7%.

In another study (Suddul and Seguin, 2023), the researchers initially designed a CNN from scratch, but this method resulted in a low accuracy of 64.6%. To improve their performance, they integrated a pre-trained EfficientNetB2 model while using TL and data augmentation for superior performance, ultimately reaching 94.5% accuracy.

Leveraging the ensemble learning method, consisting of four pre-trained models named VGG19, ResNet50, MobileNet, and AlexNet, coupled with fine-tuning and data augmentation techniques, researchers (Bu et al., 2024) achieved 96.88% accuracy, surpassing all previous studies on the Food-11 dataset.

In later research (Rokhva et al., 2024a), the researcher used MobileNetV2 pre-trained on ImageNet, with full fine-tuning and data augmentation, to achieve 92.97% accuracy with 256x256 images. They also studied other image sizes and achieved 88.2%, 77.46%, and 60,17% accuracy with 128x128, 64x64, and 32x32 images. The evaluation speed for 256-sized images, which resulted in 92.97% accuracy, was 291 images/second.

Given the outstanding performance of the EfficientNet family in Table 1, this study leveraged the pre-trained EfficientNetB7 as a backbone for feature extraction. The enriched feature map was enhanced with CBAM, an advanced attention module focusing on significant channels and spatial regions. After flattening the feature map and classifying it with 11 neurons in the FC layer, the model achieved around 96.40% accuracy with 256x256 pixel images, surpassing previous single-model studies and approaching the 96.88% accuracy obtained by (Bu et al., 2024), using an ensemble of four models.

Additionally, the evaluation speed of the current study is approximately 62 images/second with the T4 GPU. Although this number is significantly lower than the 291 image/s obtained by (Rokhva et al., 2024a) on the same hardware, since they utilized MobileNetV2 as a much simpler model, it is still very high for many practical scenarios. This is primarily because not many kitchens can serve more than 60 foods within a second. The results of all studies focused on the Food-11 dataset are summarized in Table 2.

The superior performance of this study among all single-model studies demonstrates the effectiveness of the proposed model, indicating that when the pre-trained EfficientNetB7 is fully fine-tuned and the extracted feature map enhanced with the CBAM attention module, the performance is higher than most prior research. Furthermore, while this study was not as accurate as ensemble methods, the current single model has fewer total parameters than the combination of four models. This reduction can be advantageous in inference mode, where backpropagation is absent, making the total parameter count and FLOPs critical considerations.

*Table 2 – Comprehensive comparison with related studies*

| Reference (time-based) | Employed DL model | Best reported accuracy | Additional information |
|---|---|---|---|
| (Özsert Yiğit and Özyildirim, 2018) | CaffeNet & AlexNet (separately) | 86.92% | CaffeNet + SGD: 80.51%<br>CaffeNet + Adam: 83.7%<br>AlexNet + SGD: 82.07%<br>AlexNet + Adam: 86.92%<br>Image resolution: 512x512 (High) |
| (Islam et al., 2018a) | Inception-V3 | 92.86% | TL + Data augmentation<br>Image resolution: 299x299 |
| (Islam et al., 2018b) | Classic CNN from the ground up | 74.7% | A combination of DL techniques<br>Image resolution: 224x224 |
| (Suddul and Seguin, 2023) | Classic CNN from the ground up & EfficientNetB2 (separately) | 94.5% | CNN from the scratch: 64.6%<br>EfficientNetB2: 94.5%<br>TL + Data augmentation |
| (Bu et al., 2024) | Ensemble Learning consisting four different models | 96.88% | Ensemble Learning consisting four models: VGG19, ReseNet50, MobileNet, and AlexNet.<br>All models were pre-trained but fine-tuned.<br>*Best performance among all studies* |
| (Rokhva et al., 2024a) | MobileNetV2 | 92.97% | MobileNetV2 was pre-trained and fully tuned. Data augmentation was also used.<br>Evaluation performance and speed:<br>256x256 size: 92.97% - 291 image/s<br>128x128 size: 88.20% - 452 image/s<br>64x64 size: 77.46% - 626 image/s<br>32x32 size: 60.17% - 702 image/s<br>Hardware: T4 GPU |
| Current Study | EfficientNetB7 + CBAM + FC (tailored for 11 classes) | 96.40% | Pre-trained EfficientNetB7 (backbone) + Full Fine Tuning + CBAM (enhancement) + FC (tailored for 11 classes)<br>TL + Diversified data + other DL techniques.<br>96.40% average accuracy and 62 image/s (for evaluation)<br>Image resolution: 256x256<br>*Best performance among single model studies and nearing the ensemble approach* |

## 4.2 Studying the impact of employed techniques

TL was employed to enhance and accelerate convergence, as its effectiveness is demonstrated in Figure 3. Initially, at Epoch=0, with no prior training, accuracy hovered around 10%. However, by utilizing pre-trained ImageNet weights and performing full fine-tuning, the training and evaluation accuracy surged to nearly 70% and 90% after just one epoch (617 iterations of training with 9866 images, batch size of 16).

This approach significantly reduced the need for initial excessive computations. Additionally, TL enabled the use of smaller LRs at the initial stages, facilitating the application of lower LRs in subsequent steps, resulting in lower accuracy fluctuations. Without TL, larger LRs, slower convergence, and poorer performance within the same epoch would have been expected. Furthermore, TL eliminated the necessity to calculate the mean and STD for the Food-11 dataset, as the pre-calculated ImageNet mean and STD values could be directly used across the three channels.

Regularization proved highly effective, with a small L2 regularization value of 1e-5. Although optimal evaluation accuracy was reported for each execution, according to Figure 3, accuracy in later epochs fluctuated slightly around the optimal, with no signs of overfitting or divergence, eliminating the need for early stopping. Additionally, according to Figure 3, diversifying the dataset was beneficial in preventing overfitting.

The gradual LR halving every five epochs demonstrated its effectiveness, particularly in the early epochs. As indicated in Figure 3, an initial LR of 0.01 accelerated convergence, but its positive effect lessened after five epochs due to partial network optimization, requiring a smaller LR for better convergence. Reducing the LR by 50% in the next five epochs led to a slight increase in accuracy, promoting better convergence. Even though the impact of this process diminished after 15-20 due to minimal values for LR, the process persisted, as theoretically, smaller steps result in improved convergence when approaching the minimum.

Similar to 5-fold cross-validation, the generalization approach produced consistent and logical results. As expected, the five accuracy values were similar but varied slightly due to randomness. The best and worst accuracies, 96.51% and 96.24%, respectively, being close to the average of 96.40%, further validated these expectations.

## 4.3 Confusion matrix insights and rooms for improvement

Examining Figure 6 and Figure 1 indicates that class imbalance was not the primary reason for the underperformance in certain classes. As anticipated in Section 2.3, the expectations align with the observation that similarities in patterns and color shades made recognition challenging for some classes.

Due to the frequent use of similar bright colors and patterns, in Section 2.3, we predicted that certain classes (Bread, Dairy products, Dessert, Egg, and Fried Foods) might exhibit lower classification accuracy and could be susceptible to mutual misclassification. The results shown in Figure 6 validate this regarding underperforming categories and the significant proportion of mutual misclassifications.

To provide some examples, in each training, validation, and evaluation dataset are images of hamburgers, pizzas, and pieces of bread filled with eggs or other ingredients classified as "Bread" but can significantly resemble patterns seen in the "Fried Food" category. Our observations also suggest that "Bread" is more likely to be misclassified as "Fried Food" than vice versa, a finding supported by Figure 6. Also, the frequent use of yellow and orange colors in three categories (Desserts, Dairy products, and Eggs) gave rise to mutual inaccuracies.

A similar rationale applies to the well-classified classes as well. "Rice" performed well due to its unique combination of granular texture and white color, while "Noodles-Pasta" was the only filamentous class. "Soup's" watery nature likely facilitated its accurate classification, and the "Vegetable-fruit" class benefited from distinctive colors, such as green, red, and orange, which are less common in other categories.

To improve results, we should propose several suggestions. However, we contend that concentrating on more complex models that only excel at feature extraction is unlikely to yield significant enhancements. This is primarily because neither the state-of-the-art EfficientNetB7 combined with CBAM as a single robust model (current study) nor the ensemble approach comprising four different pre-trained models (Bu et al., 2024) could exceed 97% accuracy.

Instead, we first recommend that creating more detailed classes will likely enhance the overall performance since breaking some underperformed broad classes, such as "Fried Food," into specific types, like French fries, fried chicken, and egg-fried bread, may yield better results. Moreover, even classifying well-performed classes may be justified. For example, "Seafood" can be classified into shrimp, shellfish, and fish. At the same time, "Vegetable-Fruits" could be divided into fruits and vegetables or even more granular categories.

By specifying classes, models can extract more precise and object-specific features, potentially leading to improved performance, though this should be thoroughly investigated by more research. Remarkably, though detailing classes may lead to finer categories, by doing so, some classes may have few numbers of data. Thus, increasing data quantity in each class becomes more pronounced.

The second approach can be viewed as Hard Example Mining (HEM), which presents a promising approach to address the classification challenges posed by our dataset, particularly for the visually similar classes. Given that observed mutual misclassification arises from shared features like texture and color, HEM may enhance the model's ability to distinguish between such classes.

By systematically focusing on examples the model struggles to classify (those with higher misclassification rates or ambiguous visual cues), HEM ensures that the model receives targeted training on the most challenging instances. This method can be implemented by modifying the loss function, such as "Focal Loss," which prioritizes learning complex samples over easier ones. As a result, the model will refine its understanding of subtle distinctions, such as the crispy texture that differentiates "Fried Food" from "Bread" or the fine variations between creamy "Desserts" and "Dairy products."

Integrating techniques for model interpretability and explainability can be viewed as another recommendation. Given the challenges highlighted in the confusion matrix, understanding how models arrive at their predictions can provide crucial insights for improving accuracy. Techniques such as Grad-CAM (Gradient-weighted Class Activation Mapping) can be employed to visualize the regions in an input image that significantly influence the model's decisions.

By generating heatmaps, researchers can identify whether the model focuses on relevant features, such as specific textures or colors of the food items, or is misled by extraneous background elements. This is particularly valuable for addressing misclassifications among visually similar classes, like "Bread" and "Fried Food," where identifying the model's focal points can help refine its learning process.

Last but not least, combining two or several of these factors may be helpful, specifically detailing the dataset with another approach like HEM.

**4.4 Studying the impact of hardware configuration**

The training and evaluation speeds using T4 GPU and the specified batch in Section 2.8 sizes were approximately 16 and 62 images per second, respectively, with 62 being critical for real-world applications with the model's inference mode.

To conduct more research, the performance was further evaluated using the A100 GPU, which features 40GB of GPU RAM, 83.5GB of system RAM, and 235.7GB of disk space. Under these conditions, the training and evaluation speeds reached approximately 3.80 and 8.04 batches per second, translating to 60.8 and 257.28 images per second.

Therefore, while the T4 GPU with 16GB of memory could classify nearly 60 images per second in its inference model, the A100 GPU achieved an impressive rate of approximately 250 images per second. This indicates that the A100 GPU can be well-suited for more complex scenarios if needed, despite its higher cost. Nevertheless, we conjecture that the T4 GPU remains sufficiently adequate for many practical applications.

**4.5 Policy implication**

The research on food image classification using advanced deep learning techniques presents significant implications for various sectors, including food safety, dietary management, and consumer education (Lubura et al., 2022; Moumane et al., 2023). While we did not create a specific application, such as a mobile app, we enhanced the performance while appearing quickly, offering great discussion and significant room for improvement.

The high accuracy achieved in this study can be utilized to improve food labeling and regulatory frameworks significantly. This aids in ensuring compliance with nutritional labeling standards and providing consumers with accurate and reliable information to make informed choices (Chakraborty and Aithal, 2024; Mazloumian et al., 2020).

Policymakers, in particular, have a crucial role to play in integrating AI-driven classification systems into monitoring frameworks for food establishments, thereby enhancing food safety regulations and further bolstering the accuracy and reliability of food information (Ahmadzadeh et al., 2023).

Accurate food identification, enabled by deep learning, can also significantly support dietary management initiatives (Bu et al., 2024; Rehman et al., 2017). Healthcare providers could use accurate classification models like this study to help patients track their food intake more effectively, promoting healthier eating habits and addressing obesity and related chronic diseases. Educational programs could leverage these advancements to teach about food diversity, potentially improving public health outcomes and fostering a more informed and health-conscious society.

## 5) Conclusion

Integrating artificial intelligence into modern society has led to significant advancements, particularly in the food industry and automatic kitchens, where AI-powered recognition systems offer key benefits like nutrient tracking, reducing food waste, and improving efficiency. Achieving high accuracy in food classification is crucial for the success of subsequent processes, making both accuracy and speed influential factors in productivity. However, a gap in balancing performance with fast processing motivates further research. Hence, this study utilized a dataset from Kaggle containing 16,643 images, with 60% used for training and 20% each for validation and evaluation. The dataset also exhibited difficulties regarding class imbalances, high intra-class diversity, and inter-class similarities. Our feature extraction and enhancement approach combined the pre-trained EfficientNetB7 model, renowned for its strong performance in complex tasks and state-of-the-art compound scaling method, with a Convolutional Block Attention Module (CBAM), which enhances both channel and spatial attention. Through full fine-tuning of the EfficientNetB7 backbone, coupled with regularization, dynamic learning rates, data augmentation, and effective hyperparameter tuning, our model achieved an average accuracy of 96.40% on the unseen evaluation data with no sign of overfitting or performance fluctuations. This surpassed the performance of all previous studies utilizing single DL models and approached the 96.88% accuracy achieved by an ensemble method that leveraged four different pre-trained models with fine-tuning. Despite its complexity, the model classified around 62 images per second during inference using a T4 GPU, making it suitable for numerous real-world applications. Interestingly, most misclassifications were not due to data imbalance or insufficient data but were instead linked to similarities in colors, patterns, and features between specific classes. Lastly, through an in-depth discussion of these challenges, we highlighted further areas for future improvement to enhance model performance.

## Funding

This research did not receive specific grants from funding agencies in the public, commercial, or not-for-profit sectors.

## CrediT authorship contribution statement

**Babak Teimourpour:** Validation, Supervision, Project administration, Methodology, Formal analysis. **Shayan Rokhva:** Writing – review & editing, Writing – original draft, Visualization, Validation, Software, Resources, Project administration, Methodology, Investigation, Formal analysis, Data curation, Conceptualization.

## Declaration of Generative AI and AI-assisted technologies in the writing process

While preparing this work, the authors used "Perplexity.ai", and "ChatGPT" to enhance language clarity and readability. After using this service, the authors reviewed and edited the content as needed. Therefore, they take full responsibility for the publication's content.

## Declaration of Competing Interest

The authors declare that they have no known competing financial interests or personal relationships that could have appeared to influence the work reported in this paper.